\documentclass[conference]{IEEEtran}
\IEEEoverridecommandlockouts
\usepackage{cite}
\usepackage{amsmath,amssymb,amsfonts}
\usepackage{algorithmic}
\usepackage{graphicx}
\usepackage{textcomp}
\usepackage{xcolor}
\usepackage{hyperref}
\usepackage{enumitem}
\usepackage{booktabs}
\usepackage{tabularx}
\usepackage{multirow}

\def\BibTeX{{\rm B\kern-.05em{\sc i\kern-.025em b}\kern-.08em
    T\kern-.1667em\lower.7ex\hbox{E}\kern-.125emX}}
\begin{document}


\title{Advance Simulation Method for Wheel-Terrain Interactions of Space Rovers: A Case Study on the UAE Rashid Rover{\textsuperscript{*}}
\thanks{The work was partly supported by Mohammed Bin Rashid Space Centre (MBRSC) with the aim of developing a heterogeneous swarm of semi-autonomous Lunar Rover.}
}

\author{{Ahmad Abubakar \href{https://orcid.org/0000-0002-3966-5881}{\includegraphics[scale=0.75]{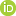}}, Ruqqayya Alhammadi\href{https://orcid.org/0000-0001-5413-8708}{\includegraphics[scale=0.75]{orcid.png}},
Yahya Zweiri \href{https://orcid.org/0000-0003-4331-7254}{\includegraphics[scale=0.75]{orcid.png}}, Lakmal Seneviratne \href{https://orcid.org/0000-0001-6405-8402}{\includegraphics[scale=0.75]{orcid.png}}
        }
\thanks{A. Abubakar, R. Alhammadi, Y. Zweiri and L. Seneviratne are with the Khalifa University Center for Autonomous and Robotic Systems (KUCARS). Y. Zweiri is also associated with Advanced Research and Innovation Center (ARIC) and the Department of Aerospace Engineering, Khalifa University. L. Seneviratne is also with the Department of Mechanical Engineering at Khalifa University, Abu Dhabi, United Arab Emirates.}
\thanks{
E-mail: \{100059792, 100041348, yahya.zweiri, lakmal.seneviratne\} @ku.ac.ae}
}



\maketitle
\begin{abstract}

A thorough analysis of wheel-terrain interaction is critical to ensure the safe and efficient operation of space rovers on extraterrestrial surfaces like the Moon or Mars. This paper presents an approach for developing and experimentally validating a virtual wheel-terrain interaction model for the UAE Rashid rover. The model aims to improve the fidelity and capability of current simulation methods for space rovers and facilitate the design, evaluation, and control of their locomotion systems. The proposed method considers various factors, such as wheel grousers properties, wheel slippage, loose soil properties, and interaction mechanics. The model's accuracy was validated through experiments on a Test-rig testbed that simulated lunar soil conditions. In specific, set of experiments were carried out to test the behaviors acted on a Grouser-Rashid rover’s wheel by the lunar soil with different slip ratios of 0, 0.25, 0.50, and 0.75. The obtained results demonstrate that the proposed simulation method provides a more accurate and realistic simulation of the wheel-terrain interaction behavior and provides insight into the overall performance of the rover. 

\end{abstract}

\begin{IEEEkeywords}
Wheel-terrain interactions, Loose soil terrain, Simulation method, Rashid rover, and Grousers.

\end{IEEEkeywords}

\section{Introduction}

The Moon has always been a source of fascination for mankind, with its mysterious surface, craters, and mountains. Over the years, several countries have sent missions to the Moon to explore its surface and study its geology \cite{b1}. The recent renewed interest in lunar exploration has led to the development of advanced technologies and space missions that are focused on exploring the Moon and establishing a permanent human presence on its surface \cite{b2}. One of the critical challenges in lunar exploration is the development of efficient and reliable unmanned vehicles that can traverse the harsh and unexplored terrain of the Moon \cite{b3}. The lunar terrain exhibits a rough and irregular topography, characterized by its coarse and uneven nature, compounded by a substantial layer of soft regolith comprising fine dust particles and fragmented rocks. The traversal of such formidable terrain poses significant impediments for rovers. Among the foremost challenges encountered by these rovers is the occurrence of slip-sinkage, an occurrence wherein the wheels of the rover undergo substantial slippage and sinkage within the loose soil \cite{b4a}. This phenomenon exerts an adverse influence on the rover's tractive capability, resulting in deviations from the desired trajectory and the inherent peril of entrenchment in the soil. For the purpose of modeling slip-sinkage, it is important to account for the wheel-terrain interaction in the context of planetary rovers navigating through deformable terrains \cite{b4}. This interaction between the wheels of space rovers and the lunar surface involves a complex interplay of physical phenomena, including soil compaction, wheel slippage, sinkage, and deformation \cite{b5}. Understanding and effectively managing these factors are essential for ensuring the successful operation of lunar rovers in such challenging environments. Therefore, it is essential to develop accurate models of the wheel-terrain interactions to design and optimize lunar rovers that can operate efficiently on the Moon's surface.

This paper introduces an approach to develop and verify the virtual simulation method specifically designed to emulate the performance of the rover's wheel on the lunar surface, as employed by the UAE Rashid rover. The proposed model incorporates the effects of wheel grousers, wheel slippage, advanced soil properties, and interaction mechanics, which involves the contact forces between the wheels and the soil, soil deformation and displacement, and resulting resistance forces acting on the wheels. The virtual simulation method was developed using the Vortex Studio, a highly realistic virtual simulator. This integration involved incorporating 3D modeling, wheel-terrain interaction mechanics, dynamics analysis, and visualization features specifically for the UAE Rashid rover. A series of experiments were performed using a single-wheel test rig, where different controlled slip ratios ranging from 0 to 1 were applied. The experimental outcomes were then compared and correlated with the results obtained from the simulation method. This comparison allowed for the refinement and optimization of the simulation model to achieve improved accuracy and precision in its predictions. The main objective of this work is to enhance the fidelity and capability of existing simulation methods for space rovers.

\subsection{Related Work}

A virtual wheel-terrain interaction model provides an effective method for analyzing traction performance without the need for physical testing \cite{b7}. This model allows researchers to predict the traction capabilities of a rover in different off-road driving scenarios, simulating various terrain conditions such as loose soil, rocky terrain, or muddy surfaces. By examining the behavior of the rover's wheels in these simulated environments, researchers can assess their performance and make informed predictions \cite{b6}. Thus, the virtual model facilitates iterative testing and optimization of the rover's wheels prior to real-world experiments, resulting in time and resource savings by reducing the necessity for numerous physical prototypes. 

Even though the virtual simulation has gotten rapid improvement in recent years \cite{b6,b8}. However, the precision of these simulation methods depends on the accuracy of the wheel-terrain interaction model employed. To improve the accuracy and potential of current simulation methods, several researchers have contributed to the development of advanced models for the interaction between the wheel of the space rovers and the terrain of the Moon environment \cite{b9}. Currently, traditional soil mechanics models such as the Bekker model and Wong-Reece model \cite{b5}, originally designed for analyzing the wheel-soil interaction mechanics of terrestrial vehicles, are commonly applied for the same purpose in planetary rovers. However, these models have limitations in their application to space rovers. The Bekker model, for instance, is capable of calculating only the static sinkage of a wheel. On the other hand, the Wong-Reece model can partially reflect the slip-sinkage generated by the longitudinal motion of soil but does not take into account the soil digging by the lugs. Even though these models are often based on simplified assumptions and do not consider the effects of various advanced soil properties, critical wheel design parameters, and operational conditions on the rover's performance, they provide a basic framework for design and evaluation \cite{b10}. Additionally, validating these models through experiments on the lunar surface is challenging and expensive \cite{b7}.

In \cite{b11}, the authors developed a virtual simulation of wheel-terrain interaction based on the Reece Pressure-sinkage model. Their focus was on a large rigid wheel on deformable terrain, and they considered variables such as the normal force, drawbar pull force, and wheel sinkage, which was also utilized in \cite{b4, b9}. To validate their model, the authors conducted experiments using a single-wheel test rig, which provided data to refine and improve the developed model. Additionally, the authors in \cite{b12} employed the use of the Beeker pressure model in conjunction with a shear stress model to simulate the interactions between a large grouser's wheel and very loose soil. They also performed experimental validation of their proposed model to ensure its accuracy before incorporating it into their virtual rover environment.

The previously discussed models have been found to have certain limitations in terms of accuracy when comparing the simulated outputs with the actual results obtained from the experimental setup. In this particular study, the primary objective was to enhance the accuracy and fidelity of the virtual simulation method. This can be achieved by incorporating additional influential factors that could minimize deviations from the experimental data and make the simulation more realistic.

\subsection{Contributions}

The aim of this study is to develop a realistic virtual simulation that effectively captures the intricacies of the wheel-terrain interaction pertaining to the UAE Rashid rover, and subsequently undergo experimental validation. Additionally, the study aims to provide valuable insights that can be utilized to improve the wheel design and control of the rover's locomotion systems. The contributions of this work are summarized as follows:

\begin{itemize}

\item{Firstly, it proposes a novel virtual simulation method for a developed wheel-terrain interaction model specifically designed for the UAE Rashid rover navigating on the target lunar environment known as the Lacus Somniorum area. By considering factors such as wheel properties, soil properties, and interaction mechanics, the model offers a more accurate representation of the complex interactions between the rover's wheels and the lunar surface.}

\item{Secondly, the paper presents the experimental validation of the proposed virtual method using a single wheel Test-rig that simulates lunar soil conditions. Through wheel-soil interaction experiments, the behaviors exhibited by a Grouser-Rashid rover's wheel when interacting with lunar soil, under different slip conditions, are thoroughly examined. The experimental results provide quantitative data that validate the accuracy and realism of the proposed model.}

\end{itemize}

\begin{figure}[htbp]
\centerline
{\includegraphics[width = \columnwidth]{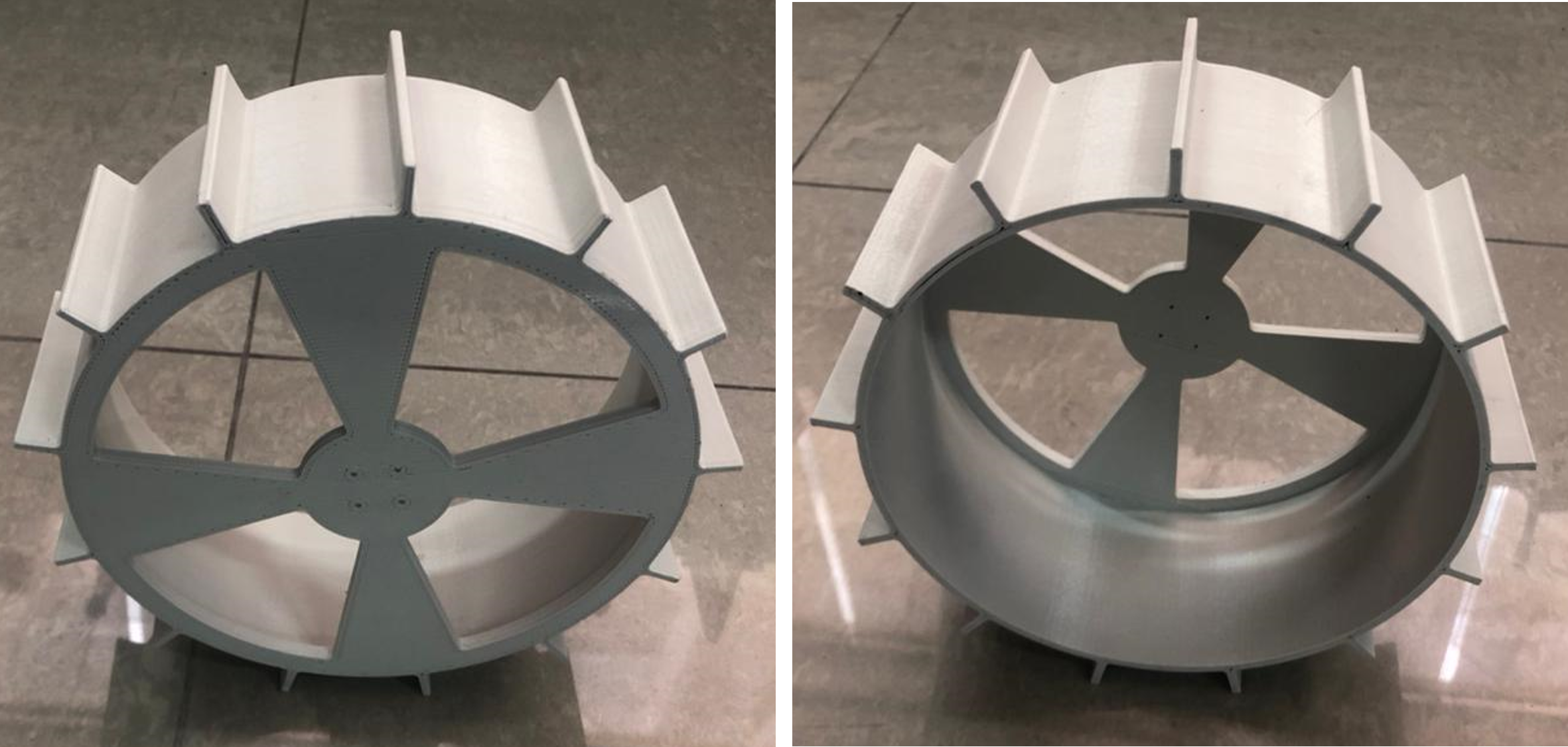}}
\caption{ Cylindrical wheel configuration with twelve fixed grousers}
\label{fig1}
\end{figure}

\section{Rover wheels}

In this research, the main emphasis is placed on the cylindrical-shaped wheels of the UAE Rashid rover. These wheels are specifically engineered to feature 12 fixed grousers or lugs, which serve the purpose of enhancing traction. Their primary function is to improve grip, particularly when traversing uneven or loose terrain, thus contributing to increased stability and control. This is particularly important for lightweight rovers that face challenges in maintaining stability and maneuverability in demanding environments. The properties of the wheel used in this research are as follows:

\begin{itemize}
    \item {Diameter: $0.1\;m$}
    \item {Width: $0.0075\;m$}
    \item {Grousers height: $0.01\;m$}
    \item {Grousers width: $0.001\;m$}
    \item {Weight: $1\;kg$}
\end{itemize}

Fig. \ref{fig1} provides a visual representation of the wheel configuration, clearly illustrating the design and arrangement of the grousers on the wheel surface.

\section{Proposed Interaction Dynamics}

The wheel-terrain interaction needs to be modeled as accurately as possible to predict more realistic and accurate interaction forces, for effective performance and safe operation of the rover. This contact interaction involves complex dynamics, including contact mechanics, deformation, and frictional effects, that significantly affect the rover's motion and stability \cite{b8}. 

Terra-mechanics models, commonly known as soft-ground tire models, have proven their capability in accurately representing the interaction between wheels and terrain in various off-road analyses. However, existing models, such as the Beeker and Wong terra-mechanics models, encounter limitations when it comes to accurately modeling the interaction of small wheels with a diameter below $500\;mm$ on extremely loose terrain \cite{b9}. To address these limitations, this research adopts the Reece pressure-sinkage model, initially developed for predicting the performance of small agricultural Unmanned Ground Vehicles (UGVs). This model provides a more suitable approach for accurately modeling the interaction of small wheels with challenging terrains. In a related study \cite{b5}, the Reece model has been modified to incorporate the influence of grousers, which are features that can reduce wheel slip on difficult terrains. The modified pressure-sinkage characteristic of homogeneous terrain is described in (1), with the effect of grousers modeled as a sinusoidal response.

\begin{equation}
p(z)=\left(c k_c^{\prime}+\gamma b k_\phi^{\prime}\right)\left(\frac{z}{b}\right)^n+A \sin (\omega t+\Phi)
\end{equation}

Here, the pressure, $p$, in the modified Reece pressure-sinkage model is a function of sinkage, $z$, and several other variables, including cohesion, $c$, dimensionless density amplitude coefficient, $k_c^{\prime}$, cohesive modulus of sinkage, $k_c$, width of the wheel, $b$, soil weight density, $\gamma$, dimensionless frictional modulus of sinkage, $k_\phi^{\prime}$, sinkage exponent, $n$, the amplitude of oscillation, $A$, and phase shift, $\Phi$. Additionally, the sinkage exponent, $n$, is calculated based on both static and dynamic sinkage, where $n=n_o+n_1 s$, with $s$ representing the dynamic relative slip. Irani et al. \cite{b9} further established the frequency of oscillation of the grousers, $\omega$, and its amplitude, $A$, as follows:

\begin{equation}
\begin{split}
\omega=\frac{\omega_w}{n_g}         \\
A &= A_\sigma+A_\gamma=\left(k_g^{\prime} \bar{\sigma}_p+k_a^{\prime} l_c d \gamma\right)
\end{split}
\end{equation}

Here, $\omega_w$ denotes the angular velocity of the wheel, $n_g$ represents the number of grousers, $A_\sigma$ signifies the amplitude of oscillation resulting from stress variation, and $A_\gamma$ denotes the amplitude resulting from density variation. Detailed descriptions of other parameters can be found in \cite{b9}.

Conversely, the traction force, which depends on the shear stress-shear strain characteristics of the soil, plays a crucial role. Among various models, the Wong shear-stress model has been demonstrated to offer improved accuracy and realism compared to other models \cite{b10}. This model establishes a relationship between shear stress, effective normal stress, and the internal friction angle of the soil. The Wong shear stress model can be expressed as the ratio of the shear force to the maximum shear force, yielding the relative shear force depicted in (3).

\begin{equation}
\frac{s}{s_{\max }}=K_r\left\{1+\left[\frac{1}{K_r(1-1 / e)}-1\right] \exp \left(1-j / K_w\right)\right\}
\end{equation}

\noindent where $s$ is the shear force, $s_{\max}$ is the maximum shear force, $K_w$ is the shear displacement where the maximum shear stress appears, and $K_r$ is the residual shear stress.

\begin{figure}[htbp]
\centerline
{\includegraphics[width = \columnwidth]{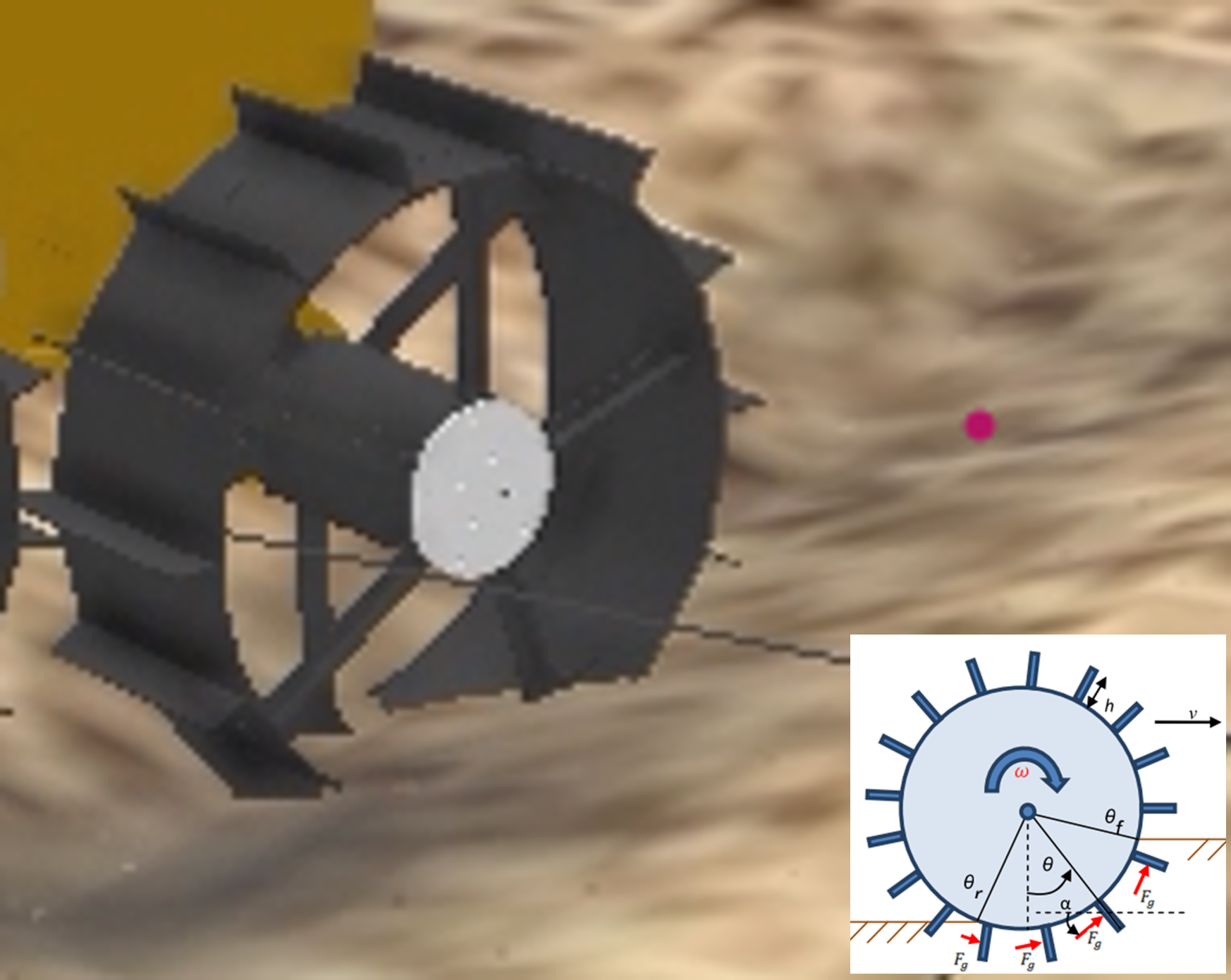}}
\caption{Evaluation of the proposed wheel-terrain interaction model on soft terrains}
\label{fig2}
\end{figure}

\section{Experimental and Simulation Set-up}

This section presents a comprehensive description of the simulation set-up and experimental set-up employed for the single-wheel Test-rig. These set-ups serve as established methodologies for precise modeling and analysis of the intricate wheel-terrain interaction, facilitating the fine-tuning of the model as required. To ensure a seamless correspondence between the simulation and the experiment, a meticulous one-to-one mapping was established between the two set-ups. Consequently, the development of a simulation set-up within a virtual environment was undertaken, accurately replicating the real-world Test-rig to maintain utmost fidelity.

\begin{figure*}[htbp]
\centerline
{\includegraphics[width = 1.0 \textwidth]{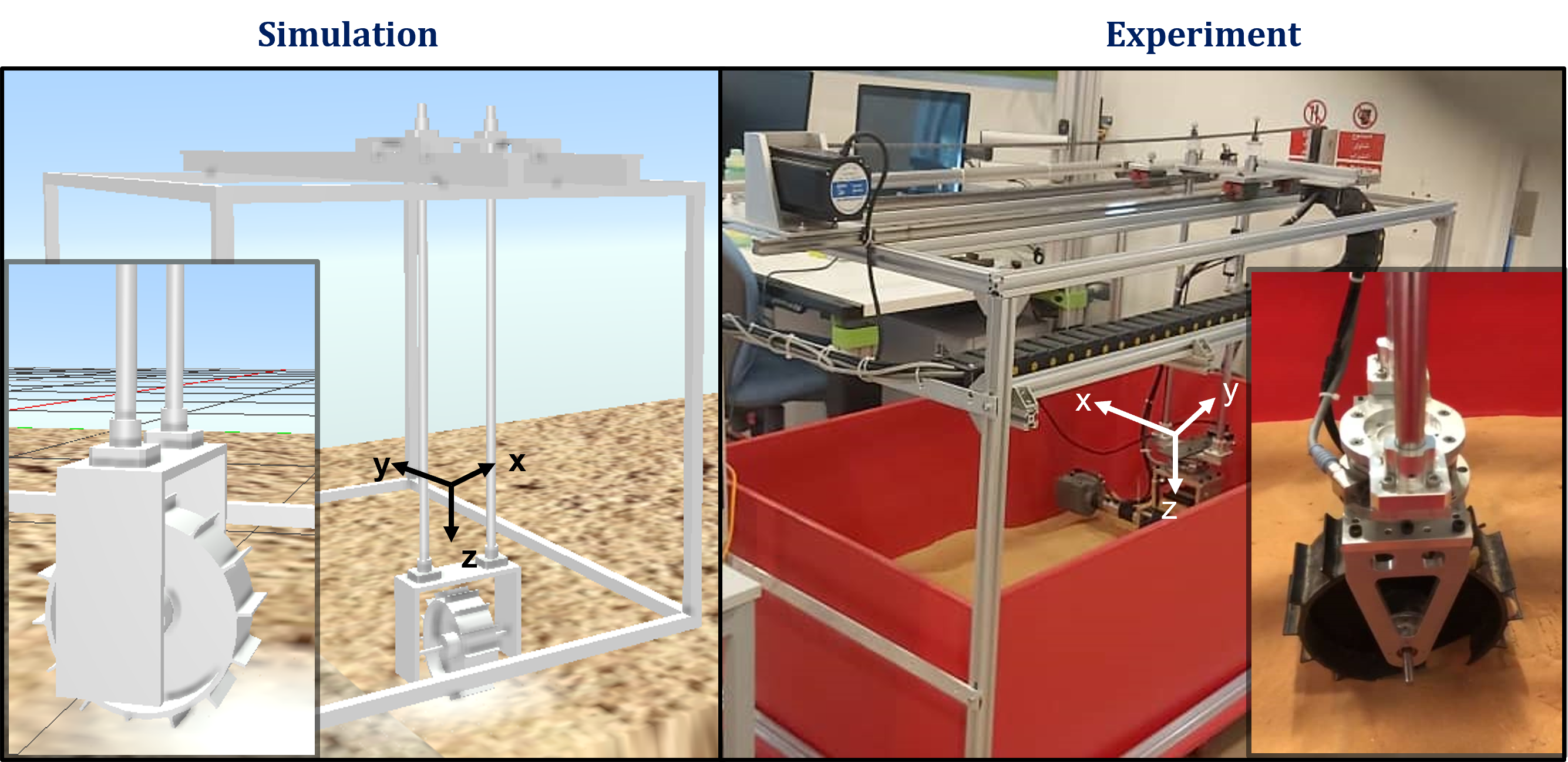}}
\caption{Combined Simulation (Left) and Experimental (Right) set-up of Test-rig visual description}
\label{fig3}
\end{figure*}

\subsection{Experimental Set-up}
\label{ss1}

A single-wheel Test-rig, depicted in Fig. 3 (Right), is utilized for the experimental phase of this project. The Test-rig was specifically developed for this research and encompasses several components, including a soil bin, soil samples, mechanical structure, safety equipment, sensors, actuators, and a control box. The Test-rig is equipped with two motors: one serves as a wheel motor that operates in closed-loop mode, while the other serves as a carriage motor for controlling the slip. This configuration allows for precise control over the wheel's operation and enables the study of different slip conditions.

The soil sample used in the Test-rig experiments is called Regolith, which was collected from the Mohammad Bin Rashid Space Center (MBRSC). The soil sample was chosen based on its anticipated properties of being very loose and deformable, specifically aiming to represent the characteristics of the Lacus Somniorum area.  This particular soil sample is utilized due to its relevance to lunar test activities conducted by the MBRSC. The properties of the soil sample were also collected to facilitate in-depth analysis and characterization. To gather measurements and data during the experiments, the physical Test-rig is equipped with various sensors, including:

\begin{itemize}
    \item {Potentiometer: This sensor is employed to measure the wheel sinkage, providing information about the depth of wheel penetration into the soil.}
    \item {ATInet Force/Torque sensor: This sensor enables simultaneous measurement of the normal force (vertical force exerted on the wheel in $z$-direction) and the Drawbar pull force (force exerted by the wheel on the Test-rig in the $y$-direction).}
\end{itemize}

These sensors enable the collection of important data related to wheel-terrain interaction forces. The measurements obtained from these sensors contribute to a comprehensive analysis of the wheel's performance and behavior during the experiments.

\subsection{Simulation Platform}
\label{ss2}

To develop the 3D simulation model, the Vortex Studio simulation environment was utilized. Vortex Studio has gained prominence as one of the foremost 3D simulators renowned for its exceptional fidelity. The development process involved the creation of a 3D model that closely matched the physical Test-rig through a one-to-one mapping approach. In more detail, the initial step was to design the 3D model of the Test-rig in Creo, a software commonly used for 3D modeling and design \cite{b14}. This ensured an accurate representation of the physical Test-rig in the virtual simulation environment. Following the creation of the 3D model, it was subsequently imported into the Vortex Studio simulation environment. The development of the simulation method in the Vortex studio consisted of various stages, including graphic gallery, assembly, and mechanics. These stages involved configuring the visual aspects of the simulation, assembling the components of the Test-rig, and defining the mechanical properties and behaviors of the simulated objects. 

We implemented the proposed wheel-terrain interaction model discussed earlier into the simulator by utilizing the pre-defined tire models and accounting for the effect of grousers/lugs available in the software's mechanism label. The soil properties, as well as the wheel properties collected from the MBRSC, were incorporated into the simulation setup. Table I provides an overview of these parameters, which were used to simulate the behavior of the soil sample. From these parameters, we can conclude that the compressibility of the soil is very high, and the internal friction and motion resistance are very low. The completed simulation setup, incorporating the configured wheel-terrain interaction model and the relevant soil and wheel properties is depicted in Fig. 3 (Left). This visualization offers a comprehensive view of the simulated environment, representing the interactions between the wheel and the soil sample.

\begin{table}[t]
\label{table1}
\caption{Summary of soil parameters}
\begin{center}
\begin{tabular}{|c|c|c|}
\hline Parameter & Value & Unit \\
\hline$k_c^{\prime}$ & 0 & - \\
\hline$k_\phi^{\prime}$ & 80 & - \\
\hline$K$ & 0.036 & $\mathrm{~m}$ \\
\hline$n$ & 1 & - \\
\hline$\gamma$ & 13734 & $\mathrm{~N} / \mathrm{m}^3$ \\
\hline$d \gamma$ & $0.1 \times \gamma$ & $\mathrm{N} / \mathrm{m}^3$ \\
\hline$\eta$ & 1.15 & - \\
\hline$\phi$ & 28 & deg \\
\hline$c$ & 0 & $\mathrm{kPa}$ \\
\hline$h_b$ & 0.01 & $\mathrm{~m}$ \\
\hline$C_f$ & 800 & $\mathrm{Ns} / \mathrm{m}$  \\
\hline$k_g^{\prime}$ & 0.06 & - \\
\hline$k_0^{\prime}$ & 0.03 & - \\
\hline
\end{tabular}
\label{tab1}
\end{center}
\vspace{-5pt}
\end{table}

\section{Experiments}\label{sec6}

In this section, a series of experimental tests were conducted to refine and validate the developed simulation method. The desired slip ratio for each experiment was calculated using the mathematical description provided in (4), as in \cite{b15}.

\begin{equation}
\text{s} = \frac{\omega_r R - V}{V}
\end{equation}

where $s$ denote the slip ratio, $\omega_r$ represents the angular velocity of the wheel motor, $R$ is the wheel radius, and $V$ denotes the linear velocity of the carriage motor.

The detailed summary of the experiments is outlined as follows:

\begin{enumerate}[label={C\arabic*}]

\item{- Wheel speed is fixed at $3 \ cm/s$, and carriage motor is set to give a slip ratio of $s = 0$.}
\item{- Wheel speed is fixed at $3 \ cm/s$, and carriage motor is set to give a slip ratio of $s = 0.25$.}
\item{- Wheel speed is fixed at $3 \ cm/s$, and carriage motor is set to give a slip ratio of $s = 0.5$.}
\item{- Wheel speed is fixed at $3 \ cm/s$, and carriage motor is set to give a slip ratio of $s = 0.75$.}
\end{enumerate}

The data obtained from these experiments encompasses the outcome variables of drawbar pull, normal force, and wheel sinkage, all of which are measured from the Test-rig. For a fair comparison, the experimental result is correlated with the simulated result from our proposed method and a theoretical model developed in \cite{b16}

\subsection{Fine-Tuning of Soil Parameters}\label{ss4}

The results obtained from the proposed simulation method and the adopted theoretical model, both configured using the soil parameters specified in Table 1, were compared with the experimental data obtained in C2, as shown in Fig. 4. The purpose of this comparison was to assess the accuracy and performance of the simulation and theoretical models, based on the collected soil parameters in replicating the experimental observations. The results of this comparison indicated that there was a lack of correlation between the simulated or theoretical values and the experimental outcomes for these variables. Specifically, when comparing the mean values, it was observed that the Drawbar pull in the experimental data was calculated to be 10.2 $N$. In contrast, the proposed simulation method yielded a value of 6.2 $N$, and the theoretical model produced a value of 5.8 $N$. Similarly, the mean value of the sinkage in the experimental data was determined to be 0.0168 $m$, while the proposed simulation method and theoretical model resulted in values of 0.0126 $m$ and 0.0118 $m$, respectively. However, the mean value of the normal force was approximately the same for all the compared methods.

The findings suggest that the observed discrepancy between the simulated values and the experimental measurements of drawbar pull and wheel sinkage can be attributed to the use of inaccurate soil parameters. Therefore, it is necessary to make further adjustments to enhance the accuracy of the models in predicting these specific variables in close alignment with the experimental data. However, it is vital to thoroughly investigate the underlying factors that contribute to these deviations. In \cite{b13}, the authors provide some insight into the most sensitive parameters that may need to be adjusted. These parameters include the internal friction angle, sinkage exponent, to name a few. By employing a trial and error approach, we were able to fine-tune the parameters and obtain more accurate parameter values for the specific soil under study. The tuned soil parameters are given in Table II.

\vspace{-5pt}

\begin{table}[htbp]
\caption{Tuned soil parameters}
\begin{center}
\begin{tabular}{|c|c|c|}
\hline Parameter & Value & Unit \\
\hline$k_c^{\prime}$ & 0.63 & - \\
\hline$k_\phi^{\prime}$ & 80 & - \\
\hline$K$ & 0.036 & $\mathrm{~m}$ \\
\hline$n$ & 1.1 & - \\
\hline$\gamma$ & 15620 & $\mathrm{~N} / \mathrm{m}^3$ \\
\hline$d \gamma$ & $0.1 \times \gamma$ & $\mathrm{N} / \mathrm{m}^3$ \\
\hline$\eta$ & 1.15 & - \\
\hline$\phi$ & 23 & deg \\
\hline$c$ & 0 & $\mathrm{kPa}$ \\
\hline$h_b$ & 0.01 & $\mathrm{~m}$ \\
\hline$C_f$ & 800 & $\mathrm{Ns} / \mathrm{m}$  \\
\hline$k_g^{\prime}$ & 0.07 & - \\
\hline$k_0^{\prime}$ & 0.035 & - \\
\hline
\end{tabular}
\label{tab2}
\end{center}
\vspace{-5pt}
\end{table}

\begin{figure}[htbp]
\centerline
{\includegraphics[width = \columnwidth]{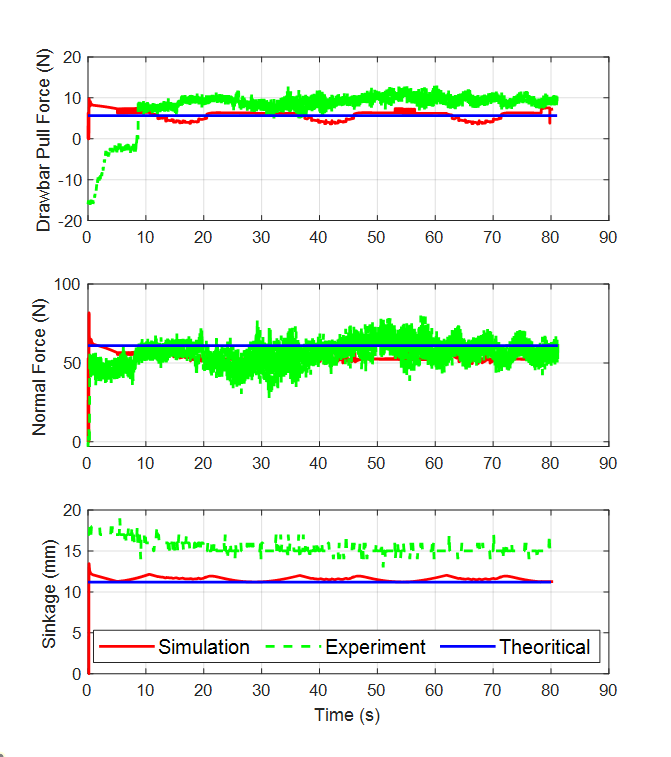}}
\caption{Performance comparison of the Test-rig experiment operating at 0.25 slip for parameter adjustment}
\label{fig}
\end{figure}

\subsection{Comparison between Simulation and Experimental Results}\label{ss5}

After successfully fine-tuning the soil parameters, the proposed simulation method and theoretical model were configured with the parameter values shown in Table 2. The experimental configurations listed above were performed, and the results for the configuration C2 are depicted in Fig. 5. Additionally, Table III provides a thorough comparison of all the experiments conducted. To facilitate data analysis, the mean values of the variable outcome have been computed and utilized for the purpose of comparison. Looking at the first configuration, C1, both the simulation method and experimental results show that the variables are approximately the same. Similarly, for configurations C2, C3, and C4, there is a close agreement between the simulation and experimental values. However, in C1, the recorded sinkage is 0.011, indicating that only the grousers of the wheel sink into the soil, which confirms the slip ratio of 0.00. On the other hand, in C2, C3, and C4, the sinkage values are recorded as 0.016, 0.0175, and 0.0208, respectively. These measurements indicate that some part of the cylindrical wheel sinks into the soil, causing some slip to occur. This observation provides evidence that supports the findings reported in \cite{b9}.

\begin{table}[]
\centering
\caption{Performance comparison for all the experiments}
\begin{tabular}{|l|ll|ll|ll|}
\hline
\multirow{2}{*}{Slip Ratio} & \multicolumn{2}{c|}{\begin{tabular}[c]{@{}c@{}}Drawbar Pull\\ (N)\end{tabular}} & \multicolumn{2}{c|}{\begin{tabular}[c]{@{}c@{}}Normal Force\\ (N)\end{tabular}} & \multicolumn{2}{c|}{\begin{tabular}[c]{@{}c@{}}Sinkage\\ (m)\end{tabular}} \\ \cline{2-7} 
                            & \multicolumn{1}{l|}{Sim}                         & Exp                          & \multicolumn{1}{l|}{Sim}                          & Exp                         & \multicolumn{1}{l|}{Sim}                       & Exp                       \\ \hline
C1 - 0.00                   & \multicolumn{1}{l|}{5.34}                        & 5.2                          & \multicolumn{1}{l|}{59}                           & 62                          & \multicolumn{1}{l|}{0.011}                     & 0.01                      \\ \hline
C2 - 0.25                   & \multicolumn{1}{l|}{9.26}                        & 10.10                        & \multicolumn{1}{l|}{58.7}                         & 62.3                        & \multicolumn{1}{l|}{0.016}                     & 0.0158                    \\ \hline
C3 - 0.50                   & \multicolumn{1}{l|}{17.2}                        & 16.77                        & \multicolumn{1}{l|}{59.1}                         & 62.1                        & \multicolumn{1}{l|}{0.0175}                    & 0.0169                    \\ \hline
C4 - 0.75                   & \multicolumn{1}{l|}{23.6}                        & 24.1                         & \multicolumn{1}{l|}{60.2}                         & 62.1                        & \multicolumn{1}{l|}{0.0208}                    & 0.0192                    \\ \hline
\end{tabular}
\vspace{-5pt}
\end{table}

\begin{figure}[htbp]
\centerline
{\includegraphics[width = \columnwidth]{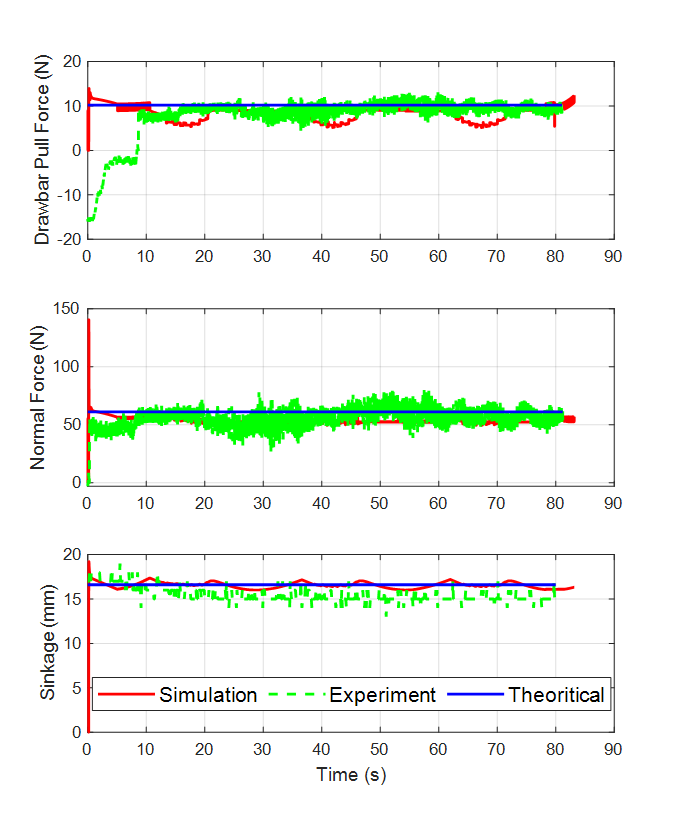}}
\caption{Performance comparison of the Test-rig experiment operating at 0.25 slip with fine-tuned parameters.}
\label{fig6}
\end{figure}

\subsection{Results and Discussion}

Based on the conducted quantitative analysis, several key observations were made regarding the slippage. These observations include:

\begin{itemize}

    \item {The magnitude of the drawbar pull is predominantly determined by the extent of slippage. As the load in the direction of movement increases, a proportional increase in the necessary drawbar pull occurs. Consequently, the slippage of the wheel exhibits a corresponding increase.}

    \item {Normal force is relatively constant with varying slippage.}

     \item {The sinkage and slippage of the wheel demonstrate a proportional correlation, wherein an increase in sinkage leads to a simultaneous increment in slippage, and vice versa.}

    \item {Although the slippage is impacted by numerous parameters, it is noteworthy that the internal friction angle, $\phi$, emerges as the most influential parameter. As the internal friction angle diminishes, denoting a softer terrain, the slippage intensifies. Furthermore, the sinkage exponent, $n$, was identified as another critical parameter exerting a notable influence on slippage. An increase in the sinkage exponent corresponds to an increase in slippage.}

\end{itemize}

The correlation of these observations with related findings in the literature  \cite{b4, b6, b9} further strengthens the conclusion that the proposed simulation method demonstrates high fidelity and accuracy in predicting the interactions between a rover's wheel and a loose soil terrain. The quantitative analysis conducted supports the effectiveness of the simulation method in capturing the influence of parameters such as the internal friction angle and sinkage exponent on slippage behavior. These findings enhance our confidence in the reliability and accuracy of the proposed simulation method.

\section{Conclusion}

This paper presents the development of a realistic simulation method for the wheel-terrain interaction of the UAE Rashid rover, aiming to enhance the fidelity and capability of current simulation methods for space rovers. The proposed model incorporates the properties of wheel grousers and loose soil to accurately simulate the interaction. The model's accuracy was validated through experimental testing using Regolith soil collected from MBRSC. The experiments involved various configurations with controlled slip ratios of 0, 0.25, 0.50, and 0.75, measuring the drawbar pull, normal force, and sinkage as outcome variables. The comparative analysis of the experimental results demonstrates that the proposed simulation method provides a more accurate and realistic representation of the wheel-terrain interaction behavior. Hence, this simulation method will be used to support the 
development of the UAE Rashid Rover virtual dynamic simulator shown in Fig. 6, operating in a Lunar environment.

Future research should focus on studying the effects of varying normal loads, as it is crucial for the advancement of space missions. The findings of this study have practical implications for optimizing the design and control of locomotion systems in lunar rovers. This contribution is significant in the development of reliable and efficient lunar space rovers, which are essential for the success of future lunar missions.

\begin{figure}[htbp]
\centerline
{\includegraphics[width = \columnwidth]{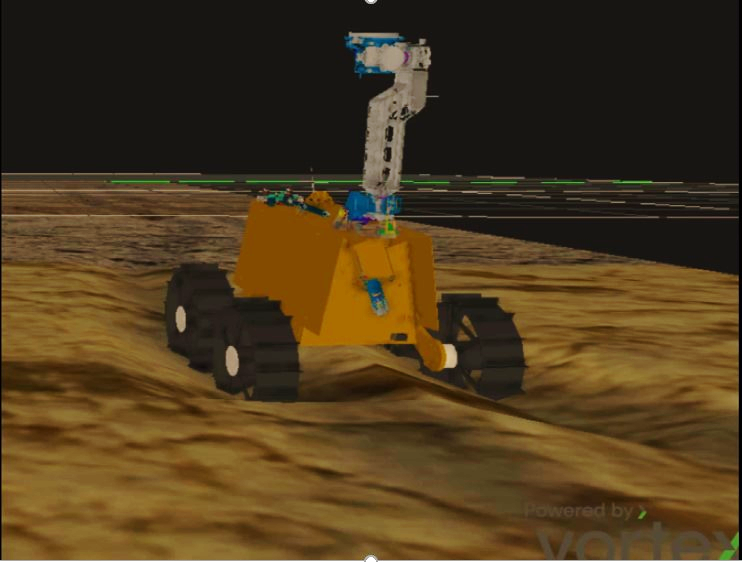}}
\caption{Visualisation of the UAE Rashid Rover operating in Lunar environment virtual simulator}
\label{fig6}
\end{figure}

\end{document}